\PassOptionsToPackage{dvipsnames}{xcolor}

\documentclass{IEEEtran} 

\usepackage{cite}
\usepackage{amsmath,amssymb,amsfonts}
\usepackage{graphicx}
\usepackage{textcomp}

\usepackage[dvipsnames]{xcolor}

\usepackage{amsthm}
\usepackage[ruled]{algorithm}
\usepackage[noend]{algorithmic}
\usepackage{alltt}
\usepackage{caption}
\usepackage{paralist}
\usepackage{nicefrac}
\usepackage{booktabs}
\usepackage{xspace}
\usepackage{mathtools}
\usepackage{marvosym}
\usepackage{wasysym}
\usepackage{verbatim}
\usepackage{subfigure}
\usepackage{hyperref}
\usepackage{multirow}
\usepackage[capitalize]{cleveref}
\usepackage[per-mode=symbol,detect-all]{siunitx}
\usepackage{graphics}
\usepackage{wrapfig}
\usepackage{enumitem}
\usepackage{float}
\usepackage{courier}
\usepackage{bm}
\usepackage{enumitem}

\usepackage{stackengine}
\usepackage[normalem]{ulem}
\usepackage{mathpartir}
\usepackage{pifont}
\usepackage{xcolor}

\theoremstyle{remark}

\usepackage{caption}
\definecolor{color0}{RGB}{228,87,46}
\definecolor{color1}{RGB}{23,190,187}
\definecolor{color2}{RGB}{255,201,20}
\definecolor{color3}{RGB}{46,40,42}
\definecolor{color4}{RGB}{118,176,65}

\definecolor{color0}{RGB}{250,121,33}
\definecolor{color1}{RGB}{254,153,32}
\definecolor{color2}{RGB}{185,164,76}
\definecolor{color3}{RGB}{86,110,61}
\definecolor{color4}{RGB}{12,71,103}

\definecolor{color0}{RGB}{228,253,225}
\definecolor{color1}{RGB}{138,203,136}
\definecolor{color2}{RGB}{100,131,129}
\definecolor{color3}{RGB}{87,87,97}
\definecolor{color4}{RGB}{255,191,70}

\definecolor{color0}{RGB}{226,59,62}
\definecolor{color1}{RGB}{243,114,44}
\definecolor{color2}{RGB}{248,150,30}
\definecolor{color3}{RGB}{249,199,79}
\definecolor{color4}{RGB}{126,179,86}
\definecolor{color5}{RGB}{67,170,139}
\definecolor{color6}{RGB}{39,125,161}
\definecolor{color7}{RGB}{21,49,60}
\definecolor{color8}{RGB}{180,215,228}

\definecolor{nnedgecolor}{RGB}{90,90,90}

\usepackage{tikz,ifthen,pgfplots}
\usetikzlibrary{arrows,trees,backgrounds,automata,shapes,decorations,plotmarks,fit,calc,positioning,shadows,chains}
\usetikzlibrary{arrows.meta,bending}
\tikzstyle{myState}=[circle,fill=blue!30!white,minimum size=17pt,inner sep=0pt]
\tikzstyle{badState}=[circle,fill=red!30!white,minimum size=17pt,inner sep=0pt]
\tikzstyle{goodState}=[circle,fill=green!30!white,minimum size=17pt,inner sep=0pt]

\tikzstyle{squareShapeBlue}=[rectangle,fill=blue!30!white,minimum size=10pt,inner sep=0pt]
\tikzstyle{squareShapeRed}=[rectangle,fill=red!30!white,minimum size=10pt,inner sep=0pt]
\tikzstyle{squareShapeGreen}=[rectangle,fill=green!30!white,minimum size=10pt,inner sep=0pt]
\tikzstyle{squareShapeCyan}=[rectangle,fill=white!30!cyan,minimum size=10pt,inner sep=0pt]
\tikzstyle{squareShapeOrange}=[rectangle,fill=orange!30!white,minimum size=10pt,inner sep=0pt]
\tikzstyle{squareShapeYellow}=[rectangle,fill=yellow!30!white,minimum size=10pt,inner sep=0pt]
\tikzstyle{squareShapeMagenta}=[rectangle,fill=magenta!30!white,minimum size=10pt,inner sep=0pt]
\tikzstyle{squareShapeWhite}=[rectangle,fill=white!30!white,minimum size=10pt,inner sep=0pt]
\tikzstyle{squareShapePink}=[rectangle,fill=pink!30!white,minimum size=10pt,inner sep=0pt]
\tikzstyle{squareShapeGray}=[rectangle,fill=gray!30!white,minimum size=10pt,inner sep=0pt]

\tikzstyle{squareResourceShapeFree}=[rectangle,fill=cyan!50!white,minimum size=9pt,inner sep=0pt]
\tikzstyle{squareResourceShapeUsed}=[rectangle,fill=magenta!50!white,minimum size=9pt,inner sep=0pt]

\tikzstyle{circleResourceShapeFree}=[circle,fill=cyan!50!white,minimum size=9pt,inner sep=0pt]
\tikzstyle{circleResourceShapeUsed}=[circle,fill=magenta!50!white,minimum size=9pt,inner sep=0pt]

\tikzstyle{every pin edge}=[<-,shorten <=1pt]
\tikzstyle{neuron}=[circle,fill=black!25,minimum size=17pt,inner sep=0pt]
\tikzstyle{input neuron}=[neuron, fill=color4]
\tikzstyle{output neuron}=[neuron, fill=color0]
\tikzstyle{hidden neuron}=[neuron, fill=color6!80]
\tikzstyle{annot} = [text width=4em, text centered]
\tikzstyle{nnedge} = [-{stealth},shorten >=0.1cm, shorten <=0.05cm,line width=0.8pt,nnedgecolor]
\usetikzlibrary{calc}

\tikzset{every picture/.style={line width=0.75pt}} 
\tikzstyle{BadSquare}=[rectangle,fill=red!30!white,minimum size=25pt,inner 
sep=0pt]
\tikzstyle{InitSquare}=[rectangle,fill=green!30!white,minimum size=25pt,inner 
sep=0pt]

\newcommand{\cmark}{\textcolor{green!80!black}{\ding{51}}}
\newcommand{\xmark}{\textcolor{red}{\ding{55}}}

\newcommand{\relu}{\text{ReLU}\xspace}

\newcommand{\sat}{\texttt{SAT}\xspace}
\newcommand{\unsat}{\texttt{UNSAT}\xspace}

\newcommand{\upperbound}{\textit{$OUT_{UPPER}$}}

\newcommand{\srlowerbound}{\textit{$SR_{\texttt{LOWER}}$}}
\newcommand{\sruperbound}{\textit{$SR_{\texttt{UPPER}}$}}
\newcommand{\smtsolver}{\texttt{DNN VERIFY }}

\newcommand{\query}{\texttt{QUERY }}
\newcommand{\packets}{\textit{P}}

\newcommand{\firstlb}{\textit{$LB_{\sat}$}}
\newcommand{\nextlb}{\textit{$LB_{\unsat}$}}

\newcommand{\firstub}{\textit{$UB_{\unsat}$}}
\newcommand{\nextub}{\textit{$UB_{\sat}$}}

\newcommand{\whirlTwo}{\textit{whiRL 2.0}}
\newcommand{\whirlOne}{\textit{whiRL 1.0}}
\newcommand{\marabou}{\textit{Marabou}\xspace}
\newcommand{\aurora}{\textit{Aurora}\xspace}
\newcommand{\pensieve}{\textit{Pensieve}\xspace}
\newcommand{\deepRM}{\textit{DeepRM}\xspace}

\usepackage{bussproofs}

\newif\ifcomments
\commentstrue

\newif\ifoutline
\outlinefalse

\newif\iflong
\longtrue

\renewcommand{\paragraph}[1]{\vspace{1mm}\noindent{\bf #1}\ }

\newcommand\blfootnote[1]{%
	\begingroup
	\renewcommand\thefootnote{}\footnote{#1}%
	\addtocounter{footnote}{-1}%
	\endgroup
}

\begin{document}

\title{Towards Scalable Verification of \\ Deep Reinforcement Learning} %

\author{
\IEEEauthorblockN{Guy Amir, Michael Schapira and Guy Katz}\\
\IEEEauthorblockA{The Hebrew University of Jerusalem, Jerusalem, Israel}

 \{guyam, schapiram, guykatz\}@cs.huji.ac.il}

\maketitle

\begin{abstract}
	\blfootnote{[*] This is the extended version
		of a paper with the same title from the FMCAD 2021 conference.
		See \url{https://fmcad.org/}}
  Deep neural networks (DNNs) have gained significant popularity in
  recent years, becoming the state of the art in a variety of
  domains. In particular, deep reinforcement learning (DRL) has
  recently been employed to train DNNs that realize control policies
  for various types of real-world systems. In this work, we present
  the \whirlTwo{} tool, which implements a new approach for verifying
  complex properties of interest for DRL systems. To demonstrate
  the benefits of \whirlTwo{}, we apply it to case studies from the
  communication networks domain that have recently been used to
  motivate formal verification of DRL systems, and which exhibit
  characteristics that are conducive for scalable verification. We
  propose techniques for performing k-induction and semi-automated
  invariant inference on such systems, and leverage these techniques for
  proving safety and liveness properties that were
  previously impossible to verify due to the scalability barriers of
  prior approaches. Furthermore, we show how our proposed techniques
  provide insights into the inner workings and the generalizability of
  DRL systems. \whirlTwo{} is publicly available online.
\end{abstract}

\section{Introduction}
\label{sec:Introduction}

In recent years, \textit{deep neural networks}
(DNNs)~\cite{GoBeCoBe16} have become highly popular due to their
ability to produce state-of-the-art results in multiple fields, e.g.,
image recognition~\cite{KrSuHi12}, text
classification~\cite{LaXuLiZh15}, game
playing~\cite{MnKaSiGrAnWiRi13}, and many
others~\cite{BoDeDwFiFlGoJaMoMuZhZhZhZi16}. DNNs used in such contexts
have been shown to successfully learn, by training on data, a model
that \emph{generalizes} to previously unseen inputs. In particular,
\emph{deep reinforcement learning} (\emph{DRL})~\cite{Li17} has been
recently used to train DNNs to learn control policies for complex
computer and networked systems, surpassing the state-of-the-art in a
variety of application domains, including database
management~\cite{ZhLiZhLiXiChXiWaChLi19}, compiler
optimization~\cite{MaJaWo20}, congestion control~\cite{LiZhChMe18,
  JaRoGoScTa19} on the Internet, routing~\cite{VaScShTa17},
compute-resource scheduling~\cite{ChXuWu17, MaAlMeKa16}, adaptive
video streaming~\cite{LeMoSuSa20, MaNeAl17}, and many more.

Despite the overwhelming success of DNNs, many safety issues
pertaining to them have been identified~\cite{GoLiZhSaYuLiWaFe20,
  SzZaSuBrErGoFe13}, demonstrating that although DNN models
potentially yield excellent performance, they also suffer from many
weaknesses. For instance, it has been shown that DNNs can be
manipulated into performing severe errors through only slight
distortions to their inputs~\cite{EnChWu20}. This phenomenon, called
\emph{adversarial perturbations}, plagues effectively all modern DNNs.

Adversarial perturbations, alongside other safety and security
vulnerabilities, have brought about a surge of interest in formally
verifying the correctness of DNNs. A plethora of approaches for DNN
verification have been proposed in recent years (e.g.,
\cite{KaBaDiJuKo21,HuKwWaWu17,GeMiDrTsChVe18,WaPeWhYaJa18}). Unfortunately,
in general, all proposed tools face significant scalability barriers,
which render them unable to verify state-of-the-art, industrial DNNs
with millions of parameters. Furthermore, even when applied to small
DNNs, these tools are often restricted to verifying simplistic
properties. The scalability challenge is further aggravated in the DRL
context, which involves \emph{sequential} DNN-informed decision
making, and so reasoning about repeated invocations of the DNN, where
the outcome of one invocation can influence the input to the DNN in
subsequent invocations. Consequently, the applicability of recently
introduced DNN verification tools to complex properties and systems of
practical interest remains extremely limited.

To begin bridging this gap, we previously introduced a tool called
\emph{\whirlOne{}}~\cite{ElKaKaSc21}, which enables verifying certain
safety and liveness properties, or identifying violations, for
practical DRL systems. We demonstrated \whirlOne{}'s usefulness by
verifying properties of interest for three systems from the
\emph{communication networking} domain. We identified such systems to
be prime candidates for verification for two main reasons: first,
state-of-the-art DNNs in this domain tend to be of moderate sizes,
which are within reach of existing verification technology; and
second, meaningful and complex specifications can be formulated and
verified because the inputs for these systems are carefully
handcrafted and reflect important semantic meaning (as opposed to raw
pixel data in computer vision applications, for example). \whirlOne{},
which combines DNN verification techniques with bounded model
checking, uses a black-box DNN verification engine as a backend, and
can thus benefit from any future improvements to DNN verification
technology. As exemplified by our promising initial results
in~\cite{ElKaKaSc21}, \whirlOne{} constituted a first step towards
enhancing the reliability of DRL systems.

Still, \whirlOne{} had severe limitations: most notably, although it
successfully generated violations of desired properties, it was
incapable of proving that properties of practical significance held
without making very strong assumptions, e.g., that runs of the
considered system terminate within a very small number of
steps. However, the executions of real-world systems are often
infinite, or finite but consisting of many steps. In such scenarios,
\whirlOne{} and other DRL verification tools are unable to prove that
most relevant properties hold.

In this work, we present \whirlTwo{}~\cite{ArtifactGitRepo} --- a
verification engine for DRL systems. \whirlTwo{} significantly extends
the capabilities of the original \whirlOne{} tool to accommodate
verifying complex properties. In particular, while \whirlOne{} was
limited to verifying basic safety properties, \whirlTwo{} utilizes
\textit{k-induction} techniques for proving both safety and liveness
properties of DRL systems. In addition, \whirlTwo{} uses
\emph{invariant inference} techniques to quickly prove properties that
could otherwise be quite difficult to verify. \whirlTwo{} also
incorporates \textit{abstraction} methods for providing some
visibility into the DRL system's operation. We demonstrate the
effectiveness of these techniques by revisiting the three case studies
involving state-of-the-art DRL systems to which \whirlOne{} has been
applied in~\cite{ElKaKaSc21}: the \aurora~\cite{JaRoGoScTa19} Internet
congestion controller, the \pensieve~\cite{MaNeAl17} adaptive video
streamer, and the \deepRM~\cite{MaAlMeKa16} compute resource
scheduler. We are able to prove various properties of these systems
that, to the best of our knowledge, were beyond the reach of prior
state-of-the-art tools, including the original \whirlOne{} tool.

The rest of this paper is organized as
follows. Section~\ref{sec:Background} covers basic background on DNNs,
DRL systems, and DNN verification. Next, in Section~\ref{sec:whirlTwo}
we present our \whirlTwo{} verification tool, and describe its
novelties and main components. We present \whirlTwo{}'s semi-automated
invariant inference in Section~\ref{sec:InvariantInference}, and
discuss the tool's implementation in
Section~\ref{sec:Implementation}. Our case studies are described in
Section~\ref{sec:CaseStudies}, followed by related work in
Section~\ref{sec:RelatedWork}. We conclude in
Section~\ref{sec:Conclusion}.

\section{Background}
\label{sec:Background}

\subsection{Deep Neural Networks and Deep Reinforcement Learning}

A deep neural network (DNN)~\cite{GoBeCoBe16} is a directed graph, where the nodes (also
called neurons) are organized in layers. In feed-forward DNNs,
data flows from the first (\emph{input}) layer, onto a
sequence of intermediate (\emph{hidden}) layers, and finally into a
final (\textit{output}) layer. The network is evaluated by assigning
values to the input layer's neurons, and then iteratively computing
the assignment of each of the hidden layers, until reaching the output
layer and returning its evaluation to the user.

More specifically, the value of each neuron in the hidden and
output layers is computed using the values of neurons in the
preceding layer. Each such layer has a \emph{type}, which determines
the exact way in which its neuron values are computed. One common
layer type is the \emph{weighted sum} layer, in which each neuron is
computed as an affine combination of the values of neurons in the
preceding layer, based on edge weights and bias values determined as
part of the DNN's training process. Another popular layer type is the
\emph{rectified linear unit} (\emph{ReLU}) layer, where each node $y$
is connected to a single node $x$ from the preceding layer, and its
value is computed by $y=\relu{}(x)=\max(0,x)$. In this paper we will
focus on weighted sum and \relu{} layers, although there exist many
additional layer types, such as \emph{max-pooling} and
\emph{hyperbolic tangent}, to which our technique may be extended.

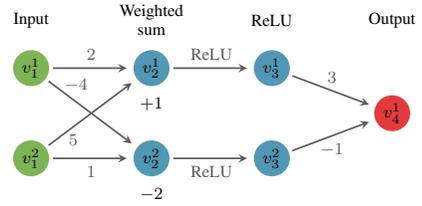
\begin{figure}[htp]
	\begin{center}
		\scalebox{0.8} {
			\def\layersep{2.0cm}
			\begin{tikzpicture}[shorten >=1pt,->,draw=black!50, node distance=\layersep,font=\footnotesize]
				
				\node[input neuron] (I-1) at (0,-1) {$v^1_1$};
				\node[input neuron] (I-2) at (0,-2.5) {$v^2_1$};
				
				\node[hidden neuron] (H-1) at (\layersep,-1) {$v^1_2$};
				\node[hidden neuron] (H-2) at (\layersep,-2.5) {$v^2_2$};
				
				\node[hidden neuron] (H-3) at (2*\layersep,-1) {$v^1_3$};
				\node[hidden neuron] (H-4) at (2*\layersep,-2.5) {$v^2_3$};
				
				\node[output neuron] at (3*\layersep, -1.75) (O-1) {$v^1_4$};
				
				\draw[nnedge] (I-1) --node[above] {$2$} (H-1);
				\draw[nnedge] (I-1) --node[above, pos=0.3] {$\ -4$} (H-2);
				\draw[nnedge] (I-2) --node[below, pos=0.3] {$5$} (H-1);
				\draw[nnedge] (I-2) --node[below] {$1$} (H-2);
				
				\draw[nnedge] (H-1) --node[above] {$\relu$} (H-3);
				\draw[nnedge] (H-2) --node[below] {$\relu$} (H-4);
				
				\draw[nnedge] (H-3) --node[above] {$3$} (O-1);
				\draw[nnedge] (H-4) --node[below] {$-1$} (O-1);

				\node[below=0.05cm of H-1] (b1) {$+1$};
				\node[below=0.05cm of H-2] (b2) {$-2$};
				
				\node[annot,above of=H-1, node distance=0.8cm] (hl1) {Weighted sum};
				\node[annot,above of=H-3, node distance=0.8cm] (hl2) {ReLU };
				\node[annot,left of=hl1] {Input };
				\node[annot,right of=hl2] {Output };
			\end{tikzpicture}
		}
		\captionsetup{size=small}
		\captionof{figure}{A toy DNN. The values above
                  the edges are weights, and the values
                  below the vertices are biases.}
		\label{fig:toyDnn}
	\end{center}
\end{figure}

Fig.~\ref{fig:toyDnn} depicts a toy DNN comprising an
input layer with two neurons, followed by a
weighted sum layer and a ReLU layer. For input
$V_1=[1, 3]^T$, the second layer's computed values are
$V_2=[18,-3]^T$. In the third layer, the \relu{} functions are applied, resulting in
$V_3=[18,0]^T$. Finally, the network's single output is $V_4=[54]$.


Formally, a DNN $N$ that receives $k$ inputs and returns $n$ outputs
is a mapping $\mathbb{R}^k\rightarrow\mathbb{R}^n$. The DNN
consists of a sequence of $m$ layers $L_1,\ldots, L_m$, where $L_1$
is the input layer and $L_m$ is the
output layer. We use $s_i$ to denote layer $L_i$'s size,
and $v_i^1,\ldots,v_i^{s_i}$ to denote $L_i$'s  individual
neurons. We refer to the column vector $[v_i^1,\ldots,v_i^{s_i}]^T$
as $V_i$. During evaluation, the input values $V_1$ are fed to the
network's input layer, and $V_2,\ldots,V_n$ are computed
iteratively.

Each weighted sum layer $L_i$ has a weight matrix $W_i$ of
dimensions $s_{i}\times s_{i-1}$ and a bias vector $B_i$ of size
$s_i$. These $W_i$ and $B_i$ are set at training time, and determine
how $V_i$ is computed: $V_i=W_i\cdot V_{i-1}+B_i$. For a ReLU 
layer $L_i$, the values of $V_i$ are computed by applying the \relu{}
to each individual neuron in its preceding layer:
$v_i^j=\relu{}(v_{i-1}^j)$. 




In \emph{deep reinforcement learning} (\emph{DRL})~\cite{Li17}, a DNN, called the \textit{agent}, learns a \emph{policy} $\pi$, which maps each possible observed
\emph{environment state} $s$ to an \emph{action} $a$. During
training, at each discrete time-step $t\in{0,1,2...}$, a
\textit{reward} $r_t$ is displayed to the agent, based on the action
$a_t$ it chose to perform after observing the environment's state at that time $s_t$. This reward is used for
tuning the agent DNN's weights. The DNNs produced using DRL fall within
the same general architecture described above; the difference lies in
the training process, which is aimed at generating a DNN that computes
a mapping $\pi$ that maximizes the
\textit{expected cumulative discounted return}
$R_t=\mathbb{E}\big[\sum_{t}\gamma^{t}\cdot r_t\big]$. The
\textit{discount factor}, $\gamma \in \big[0,1\big)$,
 controls the effect that past
decisions have on the total expected reward.


\subsection{Verification of Deep Neural Networks}
A DNN verification query typically includes a DNN $N$, a pre-condition
$P$ on $N$'s input, and a post-condition $Q$ on $N$'s
output~\cite{KaBaDiJuKo17}. The verification algorithm's goal is
to find a concrete input $x_0$ such that $P(x_0) \wedge Q(N(x_0))$
(the \sat{} case), or prove that no such $x_0$ exists (the \unsat{}
case). Typically, we use the pre-condition $P$ to express some states
of the environment that the network might encounter, and use the
post-condition $Q$ to encode the \textit{negation} of the behavior we
would like $N$ to exhibit in these states. Thus, when the verification
algorithm returns \unsat{}, this implies that the desired property
always holds. Conversely, a \sat{} result indicates that the desired
property does not always hold, and this is demonstrated by the
discovered counter-example $x_0$.

For example, observe the toy DNN in Fig.~\ref{fig:toyDnn}, and suppose
we wish to verify that the DNN's output is strictly larger than 5, for
any input, i.e., for any $x=\langle v_1^1,v_1^2\rangle$, it holds that
$N(x)=v_4^1 > 5$. This is encoded as a verification query by choosing
a pre-condition which does not restrict the input, i.e., $P=(true)$,
and by setting $Q=(v_4^1\leq 5)$, which is the \textit{negation} of
our desired property. For this verification query, a sound verifier
will return \sat{}, and a feasible counter-example such as
$x=\langle 0, -1\rangle$, which produces $v_4^1=0 \leq\ 5$. Hence,
the property does not hold for this DNN.


\medskip
\noindent
\textbf{Verifying DRL Systems.}
Beyond the general challenges of verifying DNNs (most notably, scalability), 
verifying DRL systems involves
additional challenges. These challenges stem from the fact that DRL
agents typically run within reactive systems, and are invoked
multiple times, with the inputs to each invocation usually affected by the outputs of previous
invocations. This
means that
\begin{inparaenum}[(i)]
\item the specifications for DRL systems need to account for
  multiple invocations; and
\item the scalability issue is aggravated, because the verifier
  needs to consider multiple consecutive invocations of the
  network, which is akin to considering a significantly larger DNN.
\end{inparaenum}

While attempts have been made to develop tools tailored for DRL system
verification (e.g.,~\cite{ElKaKaSc21,KaBaKaSc19,MeWaBaXuMaHu20}), two
important challenges have yet to be addressed. First, existing
verification approaches for DRL systems have focused on refuting
properties, and not on proving that they hold; and second, existing
approaches were not geared towards verifying reactive systems. As part
of the \textit{whiRL} project, we make an initial attempt at
addressing these two challenges.

\section{\whirlTwo}
\label{sec:whirlTwo}

Our contribution in this paper is the \whirlTwo{} verification tool,
which significantly extends our existing DRL verification engine,
\textit{whiRL 1.0}. The \whirlTwo{} tool allows to verify complex
queries on DRL systems, which were previously beyond our reach. Specifically,
it supports the verification of safety and liveness properties of 
DRL systems using a \textit{k-induction}-based approach. Additionally,
it incorporates \emph{invariant inference}
techniques, which facilitate the verification of complex
safety properties. \whirlTwo{}
uses an underlying verification engine as a black-box, and is hence
compatible with many existing DNN verifiers.



\medskip
\noindent
\textbf{Formalizing DRL Agents.}  DRL agents typically operate within
reactive systems: they process a (possibly infinite) sequence of
states, each representing a current snapshot of the environment
observed by the agent. Each state is obtained from its predecessor by
triggering the action outputted by the DRL agent, and allowing the
environment to react. 

In line with the formulation proposed in~\cite{ElKaKaSc21}, we
formalize the DRL verification problem by encoding the DRL system, as
well as its environment, into a transition system
$\mathcal{T}=\langle S, I, T \rangle$. Each state $s\in S$ in this
transition system is a snapshot of the current observable environment;
these states correspond to the inputs of the DNN agent. We use
$I\subseteq S$ to denote the set of initial states. The transition
relation, $T\subseteq S\times S$, is defined such that
$\langle x_i, x_j\rangle \in T$ iff the system can transition from
state $x_i$ to state $x_j$; i.e., when the DNN is presented with state
$x_i$, it selects some action, to which the environment can respond in
a way that leads the system to state $x_j$. Although the DNN is
deterministic, the environment is not necessarily so, and so $T$ need
not be deterministic. An \textit{execution} of the system is defined
as a sequence of states $x_1,\ldots,x_n$, such that $x_1 \in I$, and
for all $ 1\leq i \leq n-1$ it holds that $T(x_i,x_{i+1})$. The
process of encoding a DRL system as a transition system is supported
by \textit{whiRL 1.0}, via constructs for representing features common
to DRL systems (e.g., inputs in the form of a ``sliding window'' over
the recent history of observations)~\cite{ElKaKaSc21}.

\medskip
\noindent
\textbf{Example.}  As a running example, we focus on the
\textit{Aurora} DRL system~\cite{JaRoGoScTa19}, which implements a
congestion control policy. In today's Internet, different services
(e.g., video streaming like Netflix and Amazon, VoIP services such as
Skype)
contend over the same
network bandwidth, with aggregate demand for bandwidth often exceeding
the available supply. If Internet traffic sources do not pace the
rates at which their data is injected into the network, the network
will become congested, resulting in data being lost or delayed, and,
consequently, in bad user experience and even global Internet
outages. Congestion control is the task of determining, for each
individual Internet traffic source, how quickly its traffic should be
injected into the network at any given point in time. Congestion
control is thus a both fundamental and timely networking challenge.

Recently, researchers have proposed employing DRL for this purpose,
and presented the Aurora congestion
controller~\cite{JaRoGoScTa19}. An Aurora-controlled traffic source
uses a DNN to select the next rate at which to send traffic, based on
observations regarding the implications of its past choices of sending
rates. Specifically, Aurora's inputs are $t$ vectors
$v_{-t},\ldots,v_{-1}$, containing performance-related statistics
pertaining to the sender's most recent $t$ rate-change decisions. These incorporate
information about what fraction of sent data packets were lost
following each rate selection, how long 
it took the sent packets to reach the traffic's destination, etc. The
DNN's output determines whether the current rate should be increased,
kept steady, or decreased. Changing the sending rate can
potentially affect the environment, e.g., an increase to the rate
might lead to packet loss if the new rate exceeds network
capacity. These changes to the environment, in turn, affect the future
inputs to the DNN. See~\cite{JaRoGoScTa19} for additional details.

In the formulation of Aurora as a verification challenge
in~\cite{ElKaKaSc21}, each state, which corresponds to a possible
input to Aurora's DNN, is represented by a $t$-tuple of statistics vectors. The
state also contains the DNN's (deterministic) output for the input it
represents. This is required for defining good and bad states, as
will be discussed later. Congestion controllers are expected to
converge to ``good'' rate decisions from any starting point. Hence, we
let the set of initial states be the set of all states. Recall that
the input to the DNN represents a sliding window over $t$-long
histories of statistics vectors. Thus, for each two consecutive
states, $s_1\stackrel{T}{\rightarrow}s_2$, it holds that $s_2$ is
obtained from $s_1$ by augmenting the vectors in $s_1$ with a statistics
vector associated with the DNN's rate change at state $s_1$, and
discarding the vector in $s_1$ corresponding to the least recent of
the $t$ prior rate changes.




\medskip
\noindent
\textbf{DRL System Specifications.} Once the DRL system is
formulated as a transition system, we can specify safety and liveness
properties~\cite{ClHeVeBl18} that it should uphold. \emph{Safety
  properties} indicate that the system never displays unwanted
behavior, and these are often formulated through a predicate $P_B(s)$
that returns true iff $s\in S$ is a bad state, i.e., a state in which
the property is violated. The safety verification problem then boils
down to determining whether there is a reachable bad state in
$\mathcal{T}$~\cite{BaKa08}. \emph{Liveness properties} indicate that
the system eventually displays desirable behavior, and these are often
formulated through a predicate $P_G(s)$ that returns true iff $s\in S$
is a good state, i.e., a state in which the property is
fulfilled. Verifying a liveness property is performed by checking that
there are no infinite sequences of consecutive states in which only
finitely many of the states are good~\cite{BaKa08}. For instance, a natural safety property with respect to Aurora is that
when Aurora observes excellent network conditions (no packet loss,
close-to-minimum packet delays), as reflected by the statistics
vectors fed to the DNN, the
DRL agent does not advise to decrease the sending rate in the \textit{next
time-step}. An example of a liveness property in this setting is that if
excellent network conditions persist, Aurora should always \emph{eventually}
increase the sending rate.

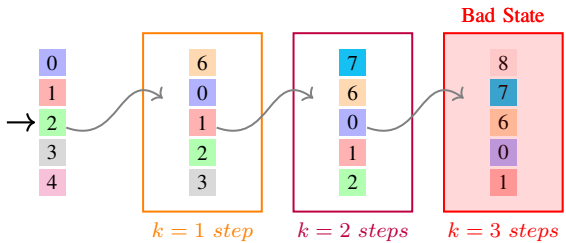
\begin{figure}[htp]
	\begin{center}
		\scalebox{1.0} {
			\def\stateSep{1.0cm}
			\def\KSepTopRow{1.2}
			\def\KSep{0.9}
			\begin{tikzpicture}[shorten >=1pt,->,draw=black!20, 
				font=\footnotesize]

			\node[squareShapeBlue] (A1) at (0,0) {0};
			\node[squareShapeRed] (A2) [below = 0.02*\stateSep of A1]{1};
			\node[squareShapeGreen] (A3) [below = 0.02*\stateSep of A2]{2};
			\node[squareShapeGray] (A4) [below = 0.02*\stateSep of A3]{3};
			\node[squareShapeMagenta] (A5) [below = 0.02*\stateSep of 
			A4]{4};	
			
			\draw[->, black, thick] ($(A3) + (-0.6cm,0)$) to (A3);

%
			
			\node[squareShapeOrange] (C1) at ($(A1) + (2cm, 0cm)$) {6};
			\node[squareShapeBlue] (C2) [below = 0.02*\stateSep of C1]{0};
			\node[squareShapeRed] (C3) [below = 0.02*\stateSep of C2]{1};
			\node[squareShapeGreen] (C4) [below = 0.02*\stateSep of C3]{2};
			\node[squareShapeGray] (C5) [below = 0.02*\stateSep of 
			C4]{3};	
			
			\node[draw,inner sep=6mm,label=below:{\color{orange}$k=1$ $step$} 
			,fit=(C2) (C4), orange] {};
			
			

			\draw[->, black, gray] (A3) to [out=-380,in=-220,looseness=1.5] 
			($(C1) - (0.5cm, 0.5cm)$);
			
			
			\node[squareShapeCyan] (D1) at ($(C1) + (2cm, 0cm)$) {7};
			\node[squareShapeOrange] (D2) [below = 0.02*\stateSep of 
			D1]{6};
			\node[squareShapeBlue] (D3) [below = 0.02*\stateSep of D2]{0};
			\node[squareShapeRed] (D4) [below = 0.02*\stateSep of D3]{1};
			\node[squareShapeGreen] (D5) [below = 0.02*\stateSep of 
			D4]{2};	
			
			\node[draw,inner sep=6mm,label=below:{\color{purple}$k=2$ $steps$} 
			,fit=(D2) (D4), purple] {};
			
			\draw[->, black, gray] (C3) to [out=-380,in=-220,looseness=1.5] 
			($(D1) - (0.5cm, 0.5cm)$);

			\node[squareShapePink] (E1) at ($(D1) + (2cm, 0cm)$) {8};
			\node[squareShapeCyan] (E2) [below = 0.02*\stateSep of 
			E1]{7};
			\node[squareShapeOrange] (E3) [below = 0.02*\stateSep of E2]{6};
			\node[squareShapeBlue] (E4) [below = 0.02*\stateSep of E3]{0};
			\node[squareShapeRed] (E5) [below = 0.02*\stateSep of 
			E4]{1};	
			
			\node[draw,inner sep=6mm,label=below:{\color{red}$k=3$ $steps$}, 
			label=above:{\color{red}Bad State}
			,fit=(E2) (E4), red, fill=red, opacity=0.15] {};
			
			\node[draw,inner sep=6mm,label=below:{}, 
			label=above:{\color{red}Bad State}
			,fit=(E2) (E4), red] {};
			
			\draw[->, black, gray] (D3) to [out=-380,in=-220,looseness=1.5] 
			($(E1) - (0.5cm, 0.5cm)$);
			
			\end{tikzpicture}
		}
		\caption{BMC searches for violations of a
		safety property. Each vector represents a state, and
                encodes the statistics  that
		Aurora observed in the past $t=5$ time-steps. The unwanted state is 
		surrounded by a red rectangle, and is reachable only after $k=3$ steps 
		from the initial state. Note that consecutive states have shared inputs 
		shifted, and each time-step sample is depicted in a different color.}
		\label{fig:bmc}
	\end{center}
\end{figure}

\medskip
\noindent
\textbf{K-Induction.} Proving that safety or liveness properties hold
(or finding counter-examples) involves traversing large transition
system graphs. For modern DRL systems, this is often infeasible, in
particular because the rich environments in which these systems
operate can react in many ways after each action taken by the agent,
resulting in high (or even infinite) out degrees for many states. In
\textit{whiRL 1.0}, this issue was addressed through the application
of \textit{bounded model checking} (BMC), an approach that explores only a
small fraction of the transition system graph, namely, states within a
$k$-step distance from an initial state. BMC can find
safety and liveness violations (if they are reachable within $k$
steps) as depicted in Fig.~\ref{fig:bmc}, but cannot prove the absence
of such violations.

In \whirlTwo{}, we address this important gap by adding the means for
proving that safety and liveness properties hold. To this end, we
employ the method of
\textit{k-induction}~\cite{ClHeVeBl18}. 

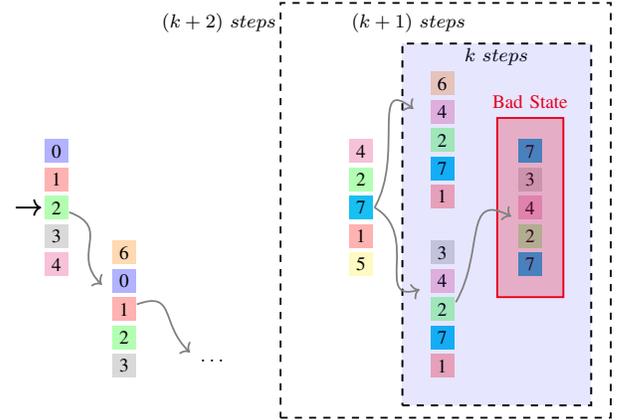
\begin{figure}[htp]
	\begin{center}
		\scalebox{0.9} {
			\def\stateSep{2.0cm}
		
			\begin{tikzpicture}[shorten >=1pt,->,draw=black!20, 
			font=\footnotesize]
			
			\node[squareShapeBlue] (A1) at (0,0) {0};
			\node[squareShapeRed] (A2) [below = 0.02*\stateSep of A1]{1};
			\node[squareShapeGreen] (A3) [below = 0.02*\stateSep of A2]{2};
			\node[squareShapeGray] (A4) [below = 0.02*\stateSep of A3]{3};
			\node[squareShapeMagenta] (A5) [below = 0.02*\stateSep of 
			A4]{4};	
			
			\draw[->, black, thick] ($(A3) + (-0.6cm,0)$) to (A3);

			\node[squareShapeOrange] (B1) at ($(A1) + (1cm, -1.5cm)$) {6};
			\node[squareShapeBlue] (B2) [below = 0.02*\stateSep of B1]{0};
			\node[squareShapeRed] (B3) [below = 0.02*\stateSep of B2]{1};
			\node[squareShapeGreen] (B4) [below = 0.02*\stateSep of B3]{2};
			\node[squareShapeGray] (B5) [below = 0.02*\stateSep of 
			B4]{3};	
			

			\draw[->, black, gray] (A3) to [out=340,in=-220,looseness=1.5] 
			($(B1) - (0.3cm, 0.5cm)$);
			
			\draw[->, black, gray] (B3) to [out=-340,in=-220,looseness=1.5] 
			($(B1) + (1cm, -1.5cm)$);

			\node at ($(B1) + (1.3cm, -1.6cm)$) {$\ldots$};
			
			%
			%
			%

			\node[squareShapeMagenta] (D1) at ($(A1) + (4.5cm, 0cm)$) {4};
			\node[squareShapeGreen] (D2) [below = 0.02*\stateSep of 
			D1]{2};
			\node[squareShapeCyan] (D3) [below = 0.02*\stateSep of D2]{7};
			\node[squareShapeRed] (D4) [below = 0.02*\stateSep of D3]{1};
			\node[squareShapeYellow] (D5) [below = 0.02*\stateSep of 
			D4]{5};

			\node[squareShapeOrange] (E1) at ($(D1) + (1.2cm, 1cm)$) {6};
			\node[squareShapeMagenta] (E2) [below = 0.02*\stateSep of 
			E1]{4};
			\node[squareShapeGreen] (E3) [below = 0.02*\stateSep of E2]{2};
			\node[squareShapeCyan] (E4) [below = 0.02*\stateSep of E3]{7};
			\node[squareShapeRed] (E5) [below = 0.02*\stateSep of 
			E4]{1};

			\node[squareShapeGray] (F1) at ($(D1) + (1.2cm, -1.5cm)$) {3};
			\node[squareShapeMagenta] (F2) [below = 0.02*\stateSep of 
			F1]{4};
			\node[squareShapeGreen] (F3) [below = 0.02*\stateSep of F2]{2};
			\node[squareShapeCyan] (F4) [below = 0.02*\stateSep of F3]{7};
			\node[squareShapeRed] (F5) [below = 0.02*\stateSep of 
			F4]{1};

			\node[squareShapeCyan] (G1) at ($(D1) + (2.5cm, 0cm)$) {7};
			\node[squareShapeGray] (G2) [below = 0.02*\stateSep of 
			G1]{3};
			\node[squareShapeMagenta] (G3) [below = 0.02*\stateSep of 
			G2]{4};
			\node[squareShapeGreen] (G4) [below = 0.02*\stateSep of G3]{2};
			\node[squareShapeCyan] (G5) [below = 0.02*\stateSep of 
			G4]{7};

			\coordinate (H3) at ($(G3) + (0.5cm, 0cm)$);
			
			\node[draw,inner sep=3mm,label=above:{\color{red}Bad State}
			,fit=(G1) (G5), red,fill=red,opacity=.25] {};
			
			\node[draw,inner sep=3mm,label=above:{}
			,fit=(G1) (G5), red] {};


			\node at ($(D1)+(+0.7cm, 1.9cm)$) {$(k+1)$ $steps$};
			
			\node at ($(E1)+(0.8cm, 0.4cm)$) {$k$ $steps$};
			
			\node at ($(D1)+(-2.1cm, 1.9cm)$) {$(k+2)$ $steps$};
			
			\node[draw,inner sep=4mm,label=above :{},  
			,fit=(E1) (F5) (H3), black, thick, dashed, 
			fill=blue,opacity=.1] {};
			
			\node[draw,inner sep=4mm,label=above :{},  
			,fit=(E1) (F5) (H3), black, thick, dashed] {};
			
			\node[draw,inner sep=10mm,label=above left:{}
			,fit=(E1) (F4) (D3)(G3), black, dashed] 
			{};

			

			\draw[->, black, gray] ($(D3) + (0.2cm, 0cm)$) to 
			[out=-300,in=-220,looseness=1.5] 
			($(E1) - (0.4cm, 0.4cm)$);

			\draw[->, black, gray] ($(D3) + (0.2cm, 0cm)$) to 
			[out=330,in=200,looseness=1.5] 
			($(F3) + (-0.3cm, 0.3cm)$);
			

			\draw[->, black, gray] ($(F3) + (0.2cm, 0.1cm)$) to 
			[out=-300,in=-210,looseness=1.5] 
			($(G3) - (0.25cm, 0.15cm)$);

		\end{tikzpicture}
		
		}
		\caption{Using k-induction to prove a safety property, 
		i.e., that the system never reaches the bad state (surrounded by a red 
		rectangle). Although there are $k$-long and $(k+1)$-long execution 
		sequences that end in the bad state, there is no such
        sequence of length $(k+2)$; and due to this and to BMC on the base 
        cases, the property holds. }
		\label{fig:k-induction}
	\end{center}
\end{figure}

Intuitively, the idea in
k-induction is to look for state sequences of length $k$, which can
start from arbitrary states in $\mathcal{T}$ (not necessarily from
initial states), and for which the property is violated. If a
violating execution exists, it must contain an indicative $k$-long
sequence of steps --- a suffix of the execution that ends in the bad
state for safety properties, or a sequence of non-good states for
liveness properties. Thus, if a verifier finds that a $k$-induction
query is \unsat{}, we know that the corresponding property holds. If,
however, it returns \sat{} with a counter-example that does not start
at an initial state, we cannot conclude whether the property holds,
and must increase $k$ in search of a conclusive answer.
Fig.~\ref{fig:k-induction} depicts a snapshot of the k-induction
process used for proving a safety property.

 More formally, following the terminology in~\cite{BaKa08}, 
 verifying $\omega$-regular 
liveness properties is reducible to checking persistence properties of the form 
\textit{"eventually 
forever $B$"}, where $B$ represents a ``bad'' state ($\exists	s $ $s.t.$ $ B=\neg 
P_G(s)$). 
Using $k$-induction in the spirit of~\cite{BiArSc02,Wa13}, we can rule out the 
existence of $k$-long sequences of bad states for a given $k$ (even ones not 
starting at an initial state). This is 
performed by formulating the following query:
\[
  \exists x_1,x_2,\ldots,x_k.
\Big(\bigwedge\limits_{i=1}^{k-1} T(x_i,x_{i+1}) \Big) \wedge
\Big(\bigwedge\limits_{i=1}^{k} \neg P_G(x_i) \Big)
\]
for increasingly large values of $k$. As soon as one such query
returns \unsat{}, we are guaranteed that the liveness property
holds. A similar encoding can be used for proving safety properties.

We note that realizing $k$-induction in our case-studies entailed
contending with challenges such as the need to encode verification
queries that capture the system-environment interaction from
\textit{any} (possibly non-initial) state.  An additional challenge
was scalability; duplicating the network to encode $k$ steps can
induce an exponential blowup in running time. \whirlTwo{} curtails the
search space by using bound tightening mechanisms, and by enforcing
certain dependencies between the inputs to the $k$ duplicate networks
encoded as part of a $k$-induction query. Specifically, these $k$
inputs typically represent the $k$ recent observations of the agent's
environment, and can be restricted by requiring them to constitute a
``sliding window'': each pair of consecutive inputs must agree on the
$k-1$ previous observations that appear in both inputs.



\begin{figure}[htp]
	 \vspace{-0.65cm}
	\begin{center}
		\scalebox{0.94} {
			\def\stateSep{1.0cm}
			\def\KSepTopRow{1.2}
			\def\KSep{0.9}
			\begin{tikzpicture}[shorten >=1pt,->,draw=black!20, 
				font=\footnotesize]
				\clip (2.5,2.5) rectangle (9.5,-2.8);
				
				\coordinate  (A0) at (4.5cm, 0cm) {};
				\node at ($(A0)+(+1.3cm, 1.7cm)$) {\texttt{verification 
				schema}\xspace};	
				
				\node at ($(A0)+(+1.3cm, -1.7cm)$) {\texttt{K++}\xspace};
				
				\draw[->, black, gray] ($(A0)+(1.25cm, -1.83cm)$) to 
				[out=-110,in=+120,looseness=2.7] 
				($(A0)+(+0.1cm, 1.8cm)$);

				\coordinate (KIndTop) at ($(G1) + (0cm, 1.0cm)$);
				\coordinate (KIndLeft) at ($(F4) + (0.5cm, +1.75)$){};	
				\coordinate (KIndRight) at ($(G2) + (0.69cm, -0.1cm)$);
				
				\node at ($(KIndTop)+(-0.055cm, 0.16cm)$) 
				{\texttt{K-Induction}\xspace};
				
				\node[draw,inner sep=4mm,label=above :{},  
				,fit=(KIndTop) (KIndLeft) (KIndRight), black, thick, dashed, 
				fill=blue,opacity=.05] {};
				
				\node[draw,inner sep=4mm,label=above :{},  
				,fit=(KIndTop) (KIndLeft) (KIndRight), black, thick, dashed] {};

				\coordinate (BmcTop) at ($(KIndTop) - (1.7cm, 0cm)$);
				\coordinate (BmcRight) at ($(KIndLeft) + (-0.9cm, 0cm)$);
				\coordinate (BmcLeft) at ($(BmcRight) + (-1.5cm, 0.61cm)$);

				\node[draw,inner sep=4mm,label=above :{},  
				,fit=(BmcTop) (BmcLeft) (BmcRight), black, thick, dashed, 
				fill=blue,opacity=.05] {};
				
				\node[draw,inner sep=4mm,label=above :{},  
				,fit=(BmcTop) (BmcLeft) (BmcRight), black, thick, dashed] {};
				
				\node at ($(BmcTop)+(-0.7cm, 0.16cm)$) {$\texttt{BMC}\xspace$};
				\coordinate (U1) at ($(G1) + (0cm, 1.0cm)$); 
				\node[draw,inner sep=10mm,label=above left:{}
				,fit=(U1) (D3) (G3), black] 
				{};

				\node[squareShapeGreen] (KIndSat) at ($(D1) + (2.5cm, -0.1cm)$) 
				{\sat};
				\node[squareShapeMagenta] (KIndUnSat) at ($(KIndSat) + (2.0cm, 
				-0.6cm)$) {\unsat};

				\draw[->, black, blue] ($(KIndSat)+(0.0cm, -0.3cm)$) to 
				[out=210,in=-380,looseness=1.5] 
				($(A0)+(+1.75cm, -1.75cm)$);
				
				\draw[->, black, green] ($(KIndTop)+(-0.1cm, -0.09cm)$) to 
				[out=210,in=-300,looseness=1.5] 
				($(KIndSat)+(0.0cm, 0.2cm)$);
				
				\draw[->, black, red] ($(KIndTop)+(-0.1cm, -0.09cm)$) to 
				[out=310,in=100,looseness=1.5] 
				($(KIndUnSat)+(0.0cm, 0.2cm)$);

				\node[squareShapeMagenta] (BmcUnSat) at ($(KIndSat) - (2.4cm, 
				0.0cm)$) 
				{\unsat};
				\node[squareShapeGreen] (BmcSat) at ($(KIndUnSat) - (6.2cm, 
				0.0cm)$) {\sat};

				\draw[->, black, blue] ($(BmcUnSat)+(0.0cm, -0.3cm)$) to 
				[out=580,in=200,looseness=1.0] 
				($(A0)+(+0.75cm, -1.75cm)$);
				
				\draw[->, black, red] ($(BmcTop)+(-0.7cm, -0.09cm)$) to 
				[out=210,in=-300,looseness=1.5] 
				($(BmcUnSat)+(0.0cm, 0.2cm)$);

				\draw[->, black, green] ($(BmcTop)+(-0.7cm, -0.09cm)$) to 
				[out=310,in=100,looseness=1.0] 
				($(BmcSat)+(0.0cm, 0.2cm)$);
				
				
				\coordinate (topKPlusLeft) at ($(BmcLeft) + (0.6cm, -2.65cm)$);
				\coordinate (topKPlusRight) at ($(topKPlusLeft) + (2.715cm, 
				0.0cm)$);				
				\coordinate (bottomKPlusLeft) at ($(topKPlusLeft) + (0.0cm, 
				0.5cm)$);
				\coordinate (bottomKPlusRight) at ($(bottomKPlusLeft) + 
				(2.55cm, 0.0cm)$);

				\coordinate (bottomLeftKPlusUpdated) at ($(bottomKPlusLeft) + 
				(0.01cm, +1.39cm) $);
				
				\coordinate (topKPlusRightUpdated) at ($(topKPlusRight) + 
				(-0.07cm, +2.45cm) $);
			
			\end{tikzpicture}

		}
		\caption{ \whirlTwo's verification schema.}
		\label{fig:verificationSchema}
	\end{center}
\end{figure}
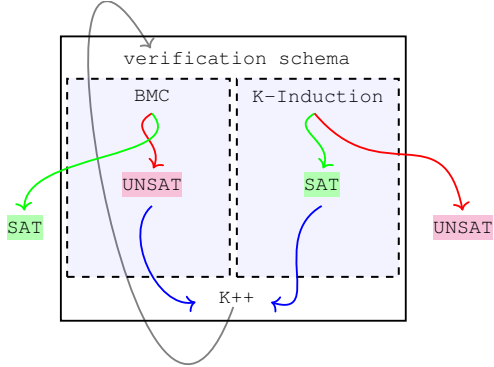

BMC and $k$-induction are related techniques; the former is geared towards
refuting a property, and the latter is geared towards proving it. In
\whirlTwo, we take a portfolio approach, as depicted in 
Fig.~\ref{fig:verificationSchema}: we alternate
between BMC and $k$-induction queries, until we:
\begin{inparaenum}[(i)]
\item refute the property (BMC returns \sat{}); or
\item prove the property ($k$-induction returns \unsat{}); or
\item hit a timeout threshold.
\end{inparaenum}
When steps 1 and 2 both fail, we increment $k$ by $1$ and repeat the process. 
Thus, although we do not know in advance whether the 
property in question holds, we hope that one of the two
techniques will either find a counter-example or prove the property.

\medskip
\noindent
\textbf{Abstraction.}
In computer networking systems, such as the Aurora 
congestion controller, the system's state is often a set of observations about the
environment. Through close inspection of our considered case-studies, we 
observe that occasionally some of
the input fields are irrelevant to the property being checked, in the sense 
that the property can be proved even when disregarding them. We thus
integrate into \whirlTwo{} \emph{abstraction}
capabilities~\cite{ClGrJhLuVe00CEGAR} --- the ability to strip off 
irrelevant
input fields, as indicated by the user, when dispatching a
verification query. The original transition system $\mathcal{T}$ is
thus changed into an abstract transition system, $\mathcal{T'}$, which
over-approximates the original one. Specifically, the states of
$\mathcal{T'}$ are symbolic, each corresponding to multiple states of
$\mathcal{T}$; and $s'_1\stackrel{T'}{\rightarrow}s'_2$ if and only if
some states $s_1$ and $s_2$, to which $s_1'$ and $s_2'$ correspond, satisfy
$s_1\stackrel{T}{\rightarrow}s_2$. If the verification engine
concludes that the property holds for $\mathcal{T'}$ (i.e., the negation of the 
property is \unsat{}), it follows that
it also holds for the original $\mathcal{T}$. However, a
counter-example for $\mathcal{T'}$ may be spurious, as it may
not be valid for $\mathcal{T}$, in which case the original 
query may need
to be solved to obtain a definite result.
 

For example, in Aurora, the DNN input
represents performance-related statistics pertaining to the $t$ most
recent rate adjustments made by the sender. In Aurora's implementation
used for our evaluation, we chose $t=10$ (as
in~\cite{JaRoGoScTa19}). In this context, abstraction might expose,
for instance, that a certain property holds regardless of what values
are assigned to the fields not relating to the $5$ most recent rate
changes, indicating that the policy is, in essence, dependent only on
the $5$ most recently observed statistics vectors.

We leverage the fact that inputs to recently-proposed computer 
 networked 
systems consist of fairly few fields with natural semantic meaning, 
thus leading to a limited number of actual combinations of 
input fields that are abstracted.

In Section~\ref{sec:CaseStudies} we demonstrate how \whirlTwo{}'s
abstraction capabilities can shed light on the inner workings of the
verified system, rendering the ``black-box'' policy learned by the DRL
system somewhat more translucent.

 
%
%

\section{Invariant Inference}
\label{sec:InvariantInference}

Verifying DRL systems is difficult, as one must often reason about transitions across many states to establish that a property holds. BMC and $k$-induction can
mitigate this issue to some extent, but sometimes this is not
enough. To further boost the scalability of \whirlTwo, we enhanced it
with semi-automated \emph{invariant inference} capabilities. 

In the context of safety verification of a transition system graph, an
\textit{invariant} can be regarded as a partition of the state space $S$ into
two disjoint sets, $S_1$ and $S_2$, such that no transition leads from one set to the other:
$s_1\in S_1\wedge s_2\in S_2\Rightarrow \langle s_1,s_2\rangle \notin T$. Invariants
are useful if we know that $I\subseteq S_1$ (all initial states
are in $S_1$) and $P_B(s)\Rightarrow s\in S_2$ (all bad states are in
$S_2$). In this case, the existence of the invariant immediately
guarantees that no bad states are reachable. Unfortunately,
discovering such useful invariants is known to be undecidable
in general, and very difficult to accomplish in practice~\cite{PaImShKaSa16}.

As part of \whirlTwo{}, we propose a heuristic for semi-automated
invariant inference, which leverages common traits of communication
networking systems. More precisely, we observe that many relevant
properties in these systems can be regarded as \textit{Boolean
  monotonic functions}; they tend to be satisfiable when the DNN's
input vectors are allowed to fluctuate extensively, but quickly become
unsatisfiable when these input vectors are restricted. Often, finding
the tipping point, i.e., the minimal input restrictions that cause the
property to shift from \sat{} to \unsat{}, constitutes an invariant
that is useful for proving other properties, and which can also render
the policy learned by the DNN more translucent to humans.

We demonstrate these notions on the Aurora congestion controller. Recall that
Aurora's output indicates whether the sending rate should be
increased, maintained, or decreased. \whirlTwo{} can
search for an invariant that translates to the range of inputs for which the DNN outputs that
the sending rate should be decreased. Such an invariant can assist in the
verification of complex properties, and provide human engineers with
comprehensible insights into the DRL system.

Technically, \whirlTwo{} allows the user to specify the output
property and mark the relevant input fields. For example, in Aurora's case,
``the sending rate should be decreased'' as the output property, and a
subset of the input statistics as the relevant fields. Then begins a binary
search on the range of the inputs in order to find the minimal
restrictions that render the verification query \unsat{}. At
each step of the binary search, we invoke a black-box verification
procedure to solve the resulting query. This allows us to locate the
tipping point up to a prescribed precision. \whirlTwo{} has built-in
\textit{templates} for input and output restrictions, which can be
regarded as different strategies for conducting the aforementioned
binary search. Each template takes into account either the DRL
system's input variables or output variables, and controls them by
adjusting their bounds; tightening them to ``push'' the query towards
the \unsat{} region. Currently, these templates include
\begin{inparaenum}[(i)]
\item
   for a fixed output, tightening or loosening the bounds of the specified input variables,
  executing binary search until the point in which the query switches
  from \sat{} to \unsat{} is discovered; and
\item
  performing a similar operation, but this time on the bounds of the
  specified output variables, while fixing the inputs according to user-specified constants.
\end{inparaenum}

\begin{wrapfigure}[12]{r}{4.5cm} 
    \vspace{-0.9cm}
	\begin{center}
		\scalebox{0.7} {
			\def\stateSep{1.0cm}
			\begin{tikzpicture}[shorten >=1pt,->,draw=black!20, 
				font=\footnotesize]

				\node[InitSquare] (GreenSquare) at (0,0) {I};
				\node[BadSquare] (RedSquare) at ($(GreenSquare) + (3.5cm, 
				3.5cm) $) {B};
				
				\coordinate (topLeft) at ($(GreenSquare) + (0cm, 3.5cm)$);
				\coordinate (bottomRight) at ($(GreenSquare) + (3.5cm, 0cm)$);
				
				\begin{scope}[on background layer]
					\draw[-, very thick, cyan, dashed] (topLeft) to 
					(bottomRight);
				\end{scope}

				\coordinate (GreenTop) at ($(topLeft) + (0cm, -1.5cm)$);
				\coordinate (GreenBottom) at ($(bottomRight) + (-1.5cm, 
				0cm)$);
				\begin{scope}[on background layer]
					\draw[-, very thick, green, dashed] (GreenTop) to 
					(GreenBottom);
				\end{scope}
				
				\coordinate (RedTop) at ($(topLeft) + (1.5cm, 0cm)$);
				\coordinate (RedBottom) at ($(bottomRight) + (0cm, 
				1.5cm)$);
				\draw[-, very thick, red, dashed] (RedTop) to 
				(RedBottom);

				\coordinate (GreenMid) at ($(GreenTop) !0.5! (GreenBottom)$);
				\coordinate (RedMid) at ($(RedTop) !0.5! (RedBottom)$);
				\coordinate (BlueMid) at ($(topLeft) !0.5! (bottomRight)$);
				
				\coordinate (betweenGreenAndBlue) at ($(GreenMid) !0.5! 
				(BlueMid)$);
				\coordinate (betweenRedAndBlue) at ($(RedMid) !0.5! 
				(BlueMid)$);
				\coordinate (topXPosition) at ($(BlueMid) + (0.15cm, 
				0.1cm)$);
				
				\draw[->, black, thick] (betweenGreenAndBlue) to 
				(betweenRedAndBlue);
				
				\node[annot, node distance=0.2cm, rotate=70] at 
				($(topXPosition) - 
				(0.02cm, 0.03cm)$)(xMarOnCrossingEdge) {\huge \xmark};

				\coordinate (GreenBegining) at (GreenSquare);
				\coordinate (GreenTop1) at ($(GreenBegining)+(0cm, 1cm)$);
				\coordinate (GreenMid1) at ($(GreenBegining)+(0.7cm, 0.7cm)$);
				\coordinate (GreenDown1) at ($(GreenBegining)+(1cm, 0cm)$);
				
				\coordinate (beforeGreenDown1) at ($(GreenDown1)-(0.7cm, 
				0.35cm)$);
				
				\begin{scope}[on background layer]
					
					\draw[->, thick] ($(GreenBegining)-(0.2cm, 0.2cm)$) to 
					[out=150,in=170,looseness=2] ($(betweenGreenAndBlue) + 
					(0cm, 
					0.15cm)$);
					
					\draw[->, thick] (GreenBegining) to (betweenGreenAndBlue);
					
					\draw [->, thick] (beforeGreenDown1) to 
					[out=350,in=310,looseness=1]($(betweenGreenAndBlue) + 
					(0.12cm, 
					-0.03cm)$);
					
					\coordinate (RedEndTop) at ($(RedSquare) - (0.45cm, 0cm)$);
					\coordinate (RedEndBottom) at ($(RedSquare) - (0cm, 
					0.45cm)$);
					
					\draw[->, thick] ($(betweenRedAndBlue) + (0.03cm, 0.03cm)$) 
					to 
					[out=420,in=240,looseness=4.3] (RedEndBottom);

				\end{scope}
				
			\end{tikzpicture}
		}
		\caption{Invariant search procedure. The initial
                  states are the green square labeled $I$, and the
                  bad states  are the red
		square labeled $B$.}
		\label{fig:invariant}
	\end{center}
\end{wrapfigure}
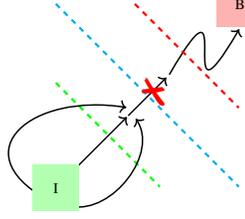

Fig.~\ref{fig:invariant} illustrates an invariant search procedure.
In this procedure, we have  a candidate invariant (the middle blue line)
 that splits the search space into two parts.
Ideally, the reachable states should all be on one side of the
partition, and the bad states on the other side. Our binary search
automatically adjusts the invariant candidate. In case an initial
invariant candidate is too strong (there are reachable states on both
sides), it is weakened, and the line is moved towards $B$. If,
however, the initial invariant candidate is too weak (there are bad
states on both sides), it is strengthened, and the line is moved
towards $I$. Both kinds of adjustments are performed by tightening or loosening
the bounds on the input or output variables.

\section{Implementation}
\label{sec:Implementation}

We implemented \whirlTwo{} as a Python framework that provides general
functionality for verifying DRL systems. \whirlTwo{} uses 
Marabou~\cite{Marabou2019}, a state-of-the-art
SMT-based~\cite{BaTi18, DeBj08, DuDe06} DNN verifier, as a backend
(although other verifiers could also be used).
\whirlTwo{} includes the following key modules, which did not exist in
\whirlOne{}:

\begin{enumerate}[leftmargin=*]

\item \textbf{K-Induction Query Verifier.}
  A module that allows the user to generate
  $k$-induction queries. The module can encode either a safety
  property or a liveness property, specified by their
  $P_B(s)$ and $P_G(s)$ predicates, respectively.
  
\item \textbf{Invariant Finder.}  A module through which a user can
  instruct \whirlTwo{} to search for an invariant. The user needs to
  provide the post-condition $Q$, and mark the variables to
  focus on. \whirlTwo{} then performs the  previously described semi-automated 
  search procedure, and returns within the specified parameters a
  range for which the invariant holds, if such a range is found.
    
\item \textbf{Input Abstraction.}  A module that allows the user to
  specify, for a given verification query, which input fields 
  should be abstracted. When abstraction is applied, \whirlTwo{} will
  either return \unsat{} (if the abstract query returns \unsat{}), or
  default to the original query if the abstract query returns a
  spurious counter-example.

\end{enumerate}
Additionally, \whirlTwo{} retains some of \textit{whiRL 1.0}'s functionality,
most notably its DNN loading interfaces and bounded model checking
capabilities. The code for \whirlTwo{}, alongside documentation and
the experiments described in the paper, are all available online under a
permissive license~\cite{ArtifactGitRepo}.

\section{Case Studies}
\label{sec:CaseStudies}

We evaluate \whirlTwo{} on three case studies of DRL systems:
the \textit{Aurora}~\cite{JaRoGoScTa19} congestion controller,
the \textit{Pensieve}~\cite{MaNeAl17} adaptive video streamer, and the
\textit{DeepRM}~\cite{MaAlMeKa16} compute resource scheduler. All three case studies, which were used to illustrate
the power of \whirlOne{} in~\cite{ElKaKaSc21}, are from
the domain of communication networks~\footnote{For a thorough formulation of 
the properties, see Appendix~\ref{appendixA}.}. We have
identified such DRL systems as highly suitable candidates for evaluating DRL 
system
verification techniques as they achieve state-of-the-art results despite being of moderate sizes, rendering verification tractable. Table~\ref{table:CaseStudySummary}
summarizes the \whirlTwo{} capabilities applied in each case
study. All experiments were conducted on an HP
EliteDesk machine with six Intel $i5-8500$ cores running at
$3.00$ GHz, and with a $32$ GB memory.

\begin{table}[t]
	\centering
	\caption{
		\whirlTwo{} features used in each case study.
	}
	\scalebox{1.0}{			
		\begin{tabular}{|c|c|c|c|}  
			\cline{2-4}
			\multicolumn{1}{c|}{} & \textbf{Aurora} & \textbf{Pensieve} & \textbf{DeepRM}  \\
			\hline
			\textit{K-Induction} & \cmark & \cmark & \xmark \\
			\hline
			\textit{Bounded Model Checking} & \cmark & \cmark & \cmark \\
			\hline
			\textit{Invariant} & \cmark & \xmark & \cmark \\
			\hline
			\textit{Abstraction} & \xmark & \cmark & \cmark \\
			\hline
		
		\end{tabular}
		   
	}
	\label{table:CaseStudySummary}
\end{table}%

\subsection{The Aurora Congestion Controller}

\textit{Aurora}~\cite{JaRoGoScTa19} is a state-of-the-art DRL 
system that acts as a congestion controller for data transmission~\cite{JaRoGoScTa19}. Aurora receives an input vector of size
$3t$, which consists of observations from the previous $t$ time-steps. 
Specifically, the input consists of $3$ distinct values representing 
performance-related statistics for each of the previous $t$ rate changes 
outputted by the DNN:
\begin{inparaenum}[(i)]
  \item \textit{latency gradient}: the derivative of latency (packet delays) 
  across time, as measured by the sender, following a change to the rate;
\item \textit{latency ratio}: the ratio of the average latency
  experienced by the sender, following a change to the rate, to the
  minimum past latency experienced. This value is never smaller than
  $1$; and
\item \textit{sending ratio}: the ratio of the rate at which packets
  are injected into the network by the sender (i.e., the sending
  rate), to the rate at which the sent packets arrive at the
  receiver. We note that the latter rate can be strictly lower than
  the former rate if the network is congested, which can lead to sent
  packets being forced to wait in in-network buffers, or
  being dropped along the way. The sending ratio is never smaller than
  $1$.
\end{inparaenum}
Intuitively, simultaneous low latency gradient, latency ratio, and 
sending ratio are indicative of excellent network conditions. Aurora has a 
single output value, which indicates whether the sending
rate should be increased (positive output),
decreased (negative output), or maintained (output is
zero). When network conditions are good (low latency, no packet loss), this in indicative of the current rate not overshooting the network bandwidth. Hence, we 
expect the 
sending rate to increase so as to take over available bandwidth. In contrast, when network conditions are poor (high latency, high 
packet loss), this is indicative of network congestion, and so we expect Aurora to decrease the rate. See~\cite{JaRoGoScTa19,ElKaKaSc21} for additional details.

In line with previous work~\cite{ElKaKaSc21,JaRoGoScTa19}, we set
$t=10$, i.e., the input size to Aurora's DNN is of size $3t=30$.
Aurora's DNN has a single hidden ReLU layer with $48$ neurons, and a
single neuron in its output layer.

\medskip
\noindent
\textbf{Proving Liveness.}  In our previous work~\cite{ElKaKaSc21}, two
liveness properties of Aurora were formulated, but could not be
verified using \textit{whiRL 1.0}. Using \whirlTwo{}, we
successfully proved that both properties from~\cite{ElKaKaSc21} always hold. Details follow.

\begin{itemize}[leftmargin=*]
  \item
\textbf{Property 1: excellent network conditions eventually imply rate increase.}
When Aurora observes a history of excellent network conditions (low latency, no 
packet 
loss), the DRL system should \textit{eventually
  increase} the sending rate, i.e., eventually output positive
values. Using \whirlTwo's $k$-induction capabilities,
we successfully proved that this property, as formulated in~\cite{ElKaKaSc21}, indeed holds for any
infinite run. The property was successfully proved, within a few seconds, for $k=2$.



\item
  \textbf{Property 2: poor network conditions eventually imply rate decrease.} 
Symmetrically to property 1, when Aurora observes a history of
poor network conditions, the DRL
system should \textit{eventually decrease} the sending rate by outputting
negative values. 
By performing $k$-induction with $k=5$, we proved that this property, as formulated in~\cite{ElKaKaSc21}, indeed holds for all infinite executions. This query took
approximately $4.5$ hours to solve.

\end{itemize}


\medskip
\noindent
\textbf{Semi-Automatic Invariance Inference.} 
Next, we used \whirlTwo{}'s invariant inference capabilities
to find invariants for proving safety properties of Aurora.


\begin{itemize}[leftmargin=*]
  \item
  \textbf{Invariant A: bounding the next-step decrease in sending rate for 
  excellent
    network conditions.}
When Aurora observes a history of excellent network conditions (low
latency, no packet loss), the DRL agent's output should be
non-negative, i.e., should not imply a decrease to the sending rate.
This safety property was shown to be violated in previous 
work~\cite{ElKaKaSc21}.
Here, we utilize \whirlTwo's invariance inference techniques to prove a 
bound on this (undesirable) next-step decrease
in sending rate, to provide visibility into the performance
of the DRL system.

\whirlTwo's method for producing the desired invariant appears in
Alg.~\ref{alg:invariantA}. The algorithm takes two user inputs: the
\textit{latency slack} $\epsilon$, and the \textit{precision} $\eta$. The
$\epsilon$ input captures the notion of ``excellent network conditions'' encoded
as inputs to the DNN: the observed latency gradient is restricted
to the range [$-\epsilon$, $\epsilon$]; and the observed latency
ratio is restricted to the range [$1$, $1+\epsilon$].
 Additionally, the 
  sending ratio is set to $1$ (indicating that sent traffic arrives at the receiver without being delayed or dropped within the network).
  The algorithm now performs a binary search over the DNN's output space 
  (leaving the prescribed input ranges for the DNN fixed).
  Specifically, the $\eta$ input specifies the desired precision:
  the output of the algorithm will be an upper bound $b$ on the DNN's output,
  such that the output $b$ is impossible, but $b+\eta$ is possible,
  given the aforementioned input restrictions. Recall that the upper bound $b$ relates to the \emph{negation} of the desired property, and so an upper bound of $b$ implies that Aurora's DNN will never decrease
the sending rate by $b$ \emph{or more} when network conditions are excellent.

  This procedure terminates within a few seconds, returning an upper
  bound on the input for which the DNN verifier returns \unsat{}. The algorithm's correctness immediately follows from the
  underlying verifier's soundness.
  
  

\begin{algorithm}
	\caption{\textit{Finding Invariant} $A$} 
	\begin{algorithmic}[1]
		\renewcommand{\algorithmicrequire}{\textbf{Input:}}
		\renewcommand{\algorithmicensure}{\textbf{Output:}}
		\REQUIRE $\epsilon$, $\eta$  
		{\color{blue}//\textit{ latency slack, precision}}
		\ENSURE \firstub{}	{\color{blue}//\textit{ worst-case output decrease
		bound}}
		\STATE \firstub{} $\leftarrow$ $ - \infty$ 
		{\color{blue}//\textit{  
		$-M$, for some large constant 
				$M$}}	
		\STATE \nextub{} $\leftarrow$ 0
		\STATE \query $\leftarrow$ \smtsolver ( $\epsilon$, output $\leq$ 0 )
		\WHILE {( $| \nextub - \firstub|  \geq \eta$ )}
		\STATE \upperbound{ } $\leftarrow$ $\frac{1}{2}$ 
		( \firstub{ } +{ } 
		\nextub{ })  
		\\ \STATE \query $\leftarrow$ \smtsolver ($\epsilon$, 
		output$\leq$ \upperbound{} )

		\STATE \textbf{if} \query is \sat{} \textbf{then} \nextub{} $\leftarrow$ \upperbound
		

                \STATE \textbf{if} \query is \unsat{} \textbf{then} \firstub{} $\leftarrow$ \upperbound

		\ENDWHILE
		\RETURN \firstub
              \end{algorithmic}
              \label{alg:invariantA}
\end{algorithm}

\item
  
  \textbf{Invariant B: inferring when Aurora fails to decrease the next-step 
  sending rate even though network
    conditions are poor.}  We now wish to characterize poor network conditions in which Aurora does not decrease its sending rate, as expected of it.
  The procedure is described in
  Alg.~\ref{alg:invariantB}. Now, the sending ratio is not fixed to 
  $1$, but is rather within the range [$1$, \packets], for a user-specified 
  \packets{} value. \packets{} represents a user-provided upper bound on ratio 
  of the rate at which packets leave the sender (i.e., the sending rate) to the 
  rate which these packets arrive at the receiver. 
  For a slack $\epsilon$, the procedure again 
  restricts the latency gradient to the range
  [$-\epsilon$, $\epsilon$] and the latency ratio to the
  range [$1$, $1+\epsilon$]. Intuitively, setting low values for $\epsilon$ while allowing sending ratios to be high corresponds to sending traffic across communication networks 
  in which in-network buffers are very shallow. In such networks, packets 
  cannot accumulate within the network, resulting in low latencies for packet 
  delivery. However, since in-network buffers are shallow, packets are dropped 
  once network bandwidth is even slightly exceeded, resulting in high sending 
  ratios when the sending rate significantly overshoots the network's capacity 
  (and many packets are lost).


The algorithm fixes the output's lower bound to be 
non-negative, and
executes a binary search on the input
sending ratio. Specifically, the algorithm returns, for any user-chosen value
\packets{}, a lower bound (\nextlb) 
such that Aurora always decreases the sending rate when its observations 
regarding past sending ratios all lie within the range 
$[\nextlb, \packets{}]$. 
\whirlTwo{} finds the invariant within a few seconds.



\begin{algorithm}
	\caption{\textit{Finding Invariant} $B$} 
	\begin{algorithmic}[1]
		\renewcommand{\algorithmicrequire}{\textbf{Input:}}
		\renewcommand{\algorithmicensure}{\textbf{Output:}}
		\REQUIRE \packets{} $\geq$ 2 {\color{blue}// \textit{upper bound on the 
		sending ratio}}
		\ENSURE  \nextlb{} {\color{blue}// \textit{worst-case sending 
		ratio bound}}
		\STATE \firstlb, \srlowerbound{} $\leftarrow$ 1 
		\STATE \nextlb, \sruperbound{} $\leftarrow$ \packets{}
		\STATE \query $\leftarrow$ \smtsolver ( $\epsilon$,  output $\geq$ 0, 
		\srlowerbound, \sruperbound{ })
		\WHILE{( \firstlb{} + 1 $ < $ \nextlb{ })}
		\STATE  \srlowerbound{} $\leftarrow$ $\frac{1}{2}$ ( \firstlb{ } +{ } 
		\nextlb{ } )  
		\STATE \query $\leftarrow$ \smtsolver ( $\epsilon$, output $\geq$ 0, 
		\srlowerbound, \sruperbound{ })
		
		

		\STATE \textbf{if} \query is \sat{} \textbf{then} \firstlb{} $\leftarrow$ \srlowerbound{}
		\STATE \textbf{if} \query is \unsat{} \textbf{then} \nextlb{} $\leftarrow$ \srlowerbound{}
		
		\ENDWHILE
 
		\RETURN \nextlb
              \end{algorithmic}
        \label{alg:invariantB}
\end{algorithm}

\end{itemize}

Observing the bounds produced by 
Alg.~\ref{alg:invariantB} yielded surprising insights regarding the
decision-making policy learned by Aurora. Specifically, to gain insight into 
what our discovered invariants reveal regarding the policies, we created 
multiple instances of Aurora agents, and trained
them all on the same training data until achieving an averaged reward value 
similar to that 
of the original Aurora controller~\cite{JaRoGoScTa19}. 
We then observed that for some of the Aurora instances, the discovered 
invariants 
depended only on the \textit{proportion} between the sending ratio's lower bound 
(\srlowerbound) and upper bound (\sruperbound), as opposed to their 
\textit{absolute} values. Specifically, for 
violating counter-examples (inputs to Aurora's DNN) produced for these instances,
the ratio between the highest and lowest past sending ratios was at least $2$, 
with lower ratios giving rise to desirable behavior by Aurora. For other 
trained instances of Aurora, violating counter-examples only depended on the 
absolute values of the bounds; e.g., Aurora always decreases the rate for 
inputs to the DNN where all sending ratios lie in the range $[1,M]$ for some 
value $M$, but not when these lie in the range $[1,M+\delta$] for some small 
$\delta$. Our findings show that policies that yield the same expected reward 
on the training set might \emph{generalize} very differently to inputs 
that lie outside this training set, and that our discovered invariants can shed 
light on the generalization strategies of different policies learned.

\subsection{The Pensieve Video Streamer}
\textit{Pensieve} is a DRL system~\cite{MaNeAl17}
for \textit{adaptive bitrate} (ABR) selection. To provide high quality of 
experience for video clients, Pensieve continuously collects statistics about 
the client's experience when downloading video chunks (e.g., was the 
video rebuffered? how long did it take to download the chunk?) to dynamically 
adapt the resolution at which the next video chunk is downloaded from the video 
server. Each video chunk represents a fixed-duration video segment (e.g., 
$4$-second-long chunks in our experiments) encoded in one of several possible 
resolutions (SD, HD, etc.), with higher resolutions corresponding to larger 
chunks, in terms of number of bits. When client-sensed network
conditions are good, we expect the ABR algorithm to decide that the
next video chunk will be downloaded in high resolution (HD); and when
they are poor, we expect a low resolution (SD) to be selected, to
avoid having the client not finish the download in time, which leads
to video rebuffering.
The input to Pensieve's DNN consists of
$(2t + M + 3)$ fields, where $t>0$ represents the number of recent video chunk downloads considered, and $M>0$ represents the number of available video resolutions. The 
input comprises:
\begin{inparaenum}[(i)]
  \item 
the \textit{bitrate} (1 field) in
which the last video chunk was downloaded;
\item 
the current \textit{video buffer
size} (1 field) of the client, reflecting the number of seconds of unwatched 
video stored at the client;
\item 
  network \textit{throughput
  measurements} for video chunks downloaded in the past $t$ time-steps ($t$ 
  fields);
\item
  \textit{download times} for the video chunks downloaded in the past $t$ 
  time-steps ($t$ 
  fields);
  \item
  \textit{resolution options} ($M$ fields) to download the next chunk;
    and
\item
  the number of \textit{remaining chunks} to be downloaded (1 field).
\end{inparaenum}
See~\cite{MaNeAl17} for a thorough exposition of
Pensieve, and~\cite{ElKaKaSc21} for a formalism of the Pensieve
verification challenge.

To maintain consistency with Pensieve's original hyper-parameters, in
our experiments $t=8$ and $M=6$. Due to the nature of an ABR
algorithm, all executions are finite (downloads finish in
finite time), and so all relevant properties are safety
properties. In previous work~\cite{ElKaKaSc21}, \whirlOne{} was
applied to check two safety properties of Pensieve:
\begin{itemize}[leftmargin=*]
\item \textbf{Property 1.} When the chunk download history represents
  \textit{excellent conditions} (short download times, large client buffer 
  size), the DRL system should \textit{increase} the 
  resolution at which chunks are requested
  before the download finishes.
\item 
  \textbf{Property 2.} When the download history represents \textit{poor network 
conditions} (long download times, small client buffer size), the DRL system 
should 
\textit{decrease} the resolution at which chunks are requested before the 
download finishes. 
\end{itemize}
While Property 1 was shown not to hold~\cite{ElKaKaSc21},
no counter-examples could previously be found for Property 2, and so it
could neither be proved nor disproved using existing tools. 
Using \whirlTwo{}, we were able to prove that Property 2 indeed holds
under certain, realistic, assumptions.\footnote{We assumed that chunks 
represent $4$-second-long video segments.
Considered chunk download times are between $4$ to $15$ seconds per chunk, 
which 
implies that downloading each chunk takes longer than consuming it.}  
To achieve this, we applied k-induction,
with $k=1$. The result returned by the verifier indicated that the bad
states are unreachable, and, hence, that the undesirable behavior cannot
occur. These verification queries took approximately 20
minutes to solve.

\subsection{The DeepRM Resource Manager}

\textit{DeepRM}~\cite{MaAlMeKa16} is a DRL-based resource manager, responsible for allocating various cluster compute resources (e.g., CPU, memory) to 
queued jobs, in order to optimize the cluster's throughput.
DeepRM receives the following as input:
\begin{inparaenum}[(i)]
  \item 
    the \textit{current resource usage} in the system;
  \item
    a \textit{queue} with up to $Q$ pending jobs waiting to be scheduled; and
  \item
    a \textit{backlog}, indicating the number of jobs waiting to be scheduled that are not yet in the queue.
\end{inparaenum}
For a fixed $Q$-sized job queue, the DeepRM controller may output one
of ($Q + 1$) possible actions: a \emph{wait} action (i.e., no resources
will be allocated at this time-step), or a \textit{schedule$_q$}
action for $1 \leq q \leq Q$, indicating that job $q$ should be
scheduled next. DeepRM's output is interpreted as a
probability distribution, assigning a certain probability to each of the
$(Q+1)$ possible actions. We refer the reader to~\cite{MaAlMeKa16} for a 
thorough exposition of DeepRM, and to~\cite{ElKaKaSc21} for a formalism of the 
DeepRM verification challenge. 

In our case study, as in~\cite{ElKaKaSc21}, we used a DeepRM system trained with $R=2$
resources: \textit{CPU} and \textit{memory units}, and a job queue of
size $Q=5$. Overall system resources consist of $10$ CPUs and $10$ memory 
units. We considered two kinds of jobs:
\textit{small} jobs, which require $1$ CPU and $1$ memory unit for a
single time-step, and \textit{large} jobs, which require $10$ CPUs and $10$ memory units, for
$t=20$ time-steps.

Previous work~\cite{ElKaKaSc21} considered the following safety properties for 
DeepRM:
\begin{itemize}[leftmargin=*]
\item \textbf{Property 1.} When  all resources are fully available, 
  and the queue is filled with \textit{small} jobs, DeepRM should 
  never assign
  the highest probability to the \emph{wait} action.
    
\item \textbf{Property 2.} When no resources are available, 
and the
  queue is filled with \textit{small} jobs, DeepRM should assign the highest
  probability to the \emph{wait} action.
  
\item \textbf{Property 3.} When no resources are available, and the
  queue is filled with \textit{large} jobs,
  DeepRM should assign the highest
  probability to the \emph{wait} action.
  
\end{itemize}
Using \textit{whiRL 1.0}, it was shown~\cite{ElKaKaSc21} that Property $1$ 
holds,
and that there exist counter-examples for Properties $2$ and  $3$.
However, by using \whirlTwo{} we were able to prove (within a few
seconds) a stronger property that, in fact, generalizes properties
$1$, $2$ and $3$. By applying \whirlTwo{}'s
abstraction capabilities to both the inputs indicating resource
utilization and the output indicating the recommended action,
we proved that for \textit{any} resource
utilization level, when the queue is filled with identical jobs, the DRL
system's output assigns a higher probability to
\textit{schedule$_2$} than to \emph{wait}. This
immediately proves Property $1$, and implies that Properties $2$ and
$3$ cannot hold.

This finding sheds new light on previous results, and enhances 
our understanding of DeepRM:
\begin{inparaenum}[(i)]
\item the three original
properties do not depend on the current resource utilization.
Rather, due to the DRL system learning a suboptimal policy,
it is biased towards scheduling a specific job (job \#$2$), and may fail to select
\emph{wait} when appropriate; and
\item  
the counter-examples found for Properties $2$ and 
 $3$ are not outliers, but rather the general case. Indeed, we
were able to use \whirlTwo{} to prove that the inverses of both these properties 
always hold.
\end{inparaenum}
These results demonstrate that, beyond proving or disproving specific
properties, \whirlTwo{} can shed light on the
policy learned by the DRL system, and expose problematic issues.

\section{Related Work}
\label{sec:RelatedWork}
Due to the increasing use of DNNs, many DNN verification tools have
been proposed in recent years;  
some are SMT-based
(e.g.,~\cite{KaBaDiJuKo17, PuTa12, Marabou2019, KuKaGoJuBaKo18}), 
whereas others use different verification strategies, 
such as \textit{abstract 
interpretation}~\cite{SiGeMiPuVe18,WeZhChSoHsDaBoDh18,ZhWeChHsDa18},
\textit{mixed integer linear programming} (MILP)~\cite{TjXiTe17}, and many
others. Recently, these approaches were extended to verify
systems with multi-step executions, such as Recurrent Neural
Networks (RNNs)~\cite{ZhShGuGuLeNa20, JaBaKa20} or hybrid
systems~\cite{SuKhSh19}.

In our evaluation of \whirlTwo{}, we used
\marabou{}~\cite{Marabou2019,WuOzZeIrJuGoFoKaPaBa20} as a black-box
DNN verifier. To date, Marabou has mostly been applied for solving
adversarial robustness queries~\cite{KaBaDiJuKo17Fvav, CaKaBaDi17,
  GoKaPaBa18, AmWuBaKa21}, and our work demonstrates that it is
also applicable in the field of computer and networked systems. Marabou
affords additional features, such as built-in
abstraction~\cite{ElGoKa20}, simplification~\cite{GoFeMaBaKa20,LaKa21},
repair~\cite{GoAdKeKa20} and optimization~\cite{StWuZeJuKaBaKo20}
techniques, which could also be applied to our case studies.

In addition to general DNN verification engines, methods
have been devised to formally verify safety properties of DRL
systems, which are the subject matter of this work. Such approaches
include \textit{shield synthesis}~\cite{KoLoJaBl20}, and combining the
verification process with \textit{verified runtime
  monitoring}~\cite{FuPa18}. Other methods focus on finding
adversarial attacks that pertain specifically to DRL agents, e.g., by using
MILP~\cite{DeCaNa21}.

In addition to the \textit{whiRL} project, other approaches have been proposed
for verifying DRL systems in 
the domain of communication networks. These include, e.g., 
\textit{Verily}~\cite{KaBaKaSc19} and 
\textit{Metis}~\cite{MeWaBaXuMaHu20}. Importantly, however, our focus is on verifying
(as opposed to only refuting) various safety and liveness properties
of these systems. To the best of our knowledge, this lies beyond the
grasp of other existing tools.

\section{Conclusion}
\label{sec:Conclusion}

DRL systems provide excellent performance in multiple settings,
but suffer from severe vulnerabilities. Several
verification tools have been developed to mitigate this concern, but these 
mostly refute, as opposed to prove, safety and liveness
properties of interest. In this
work, we presented \whirlTwo{ }--- a novel verification engine that supports 
proving both safety and liveness properties of DRL systems. \whirlTwo{} 
accomplishes this through semi-automatic invariance inference, alongside 
techniques 
such as k-induction and query abstraction. We
demonstrated our tool's capabilities through three case studies from the
communication networks domain. In addition, we demonstrated how
\whirlTwo{} can provide insights into the inner workings of these systems,
uncovering weaknesses that would otherwise remain unnoticed.

In the future, we plan to enhance our tool's scalability by using
improved search heuristics. Also, we intend to enrich the
semi-automatic invariant inference templates to support searching for
more complex invariants.

\medskip
\noindent
\textbf{Acknowledgements.}  
We thank Nathan Jay, Tomer Eliyahu and the anonymous reviewers for their 
contributions to this project.  
The project was partially supported by the Israel Science Foundation (grant 
number 683/18), the Binational Science Foundation (grant numbers
2017662 and 2019798),
and the Center for 
Interdisciplinary Data Science Research at The Hebrew University of Jerusalem.

{
\bibliographystyle{abbrv}
\bibliography{references}
}

\newpage
\newpage
\begin{appendices}

\section{Formulation of the Properties Verified}
\label{appendixA}

	This appendix describes the formal definition 
	of each of the properties proven in the paper. 
	All properties were 
	proven to hold. For violated properties, please see~\cite{ElKaKaSc21}.
	
	\subsection{\aurora Properties}

	\begin{itemize}[leftmargin=*]
		\item
		\textbf{Property 1:} excellent network conditions eventually imply rate 
		increase.
		
		For all previous time-steps ($\forall$ $t' \in \{0,1,...,t=9\}$):
		
		\subitem - latency gradient ($t$ inputs): 
		\subitem 
		$input_{3t'} \in 
		[-\epsilon, 
		\epsilon]$ 
		
		\subitem - latency ratio ($t$ inputs):  
		\subitem 
		$input_{3t'+1} \in [1, 
		1+\epsilon]$ 
		
		\subitem - sending ratio ($t$ inputs):  
		\subitem 
		$input_{3t'+2} = 1$
		
		\subitem - suggested rate factor ($1$ output): 
		\subitem 
		$output > 0$ \\
		
		\item
		\textbf{Property 2:} poor network conditions eventually imply rate 
		decrease. 
		
		For all previous time-steps ($\forall$ $t' \in \{0,1,...,t=9\}$):
		
		\subitem - latency gradient ($t$ inputs): 
		\subitem 
		$input_{3t'} \in 
		[-\epsilon, 
		\epsilon]$ 
		
		\subitem - latency ratio ($t$ inputs): 
		\subitem 
		$input_{3t'+1} \in [1, 
		1+\epsilon]$ 
		
		\subitem - sending ratio ($t$ inputs): 
		\subitem 
		$input_{3t'+2} \geq 2$
		
		\subitem - suggested rate factor ($1$ output): 
		\subitem 
		$output < 0$
	\end{itemize}

	\subsection{\pensieve Properties}
	
	\begin{itemize}[leftmargin=*]
		
		\item 
		\textbf{Property 2.} When the download history represents \textit{poor 
			network conditions} (long download times, small client buffer 
			size), the 
		DRL system 	should \textit{decrease} the resolution at which chunks are 
		requested before the download finishes.

		For $M$ resolution 
		options~\footnote{Resolutions $\{res_{0},...,res_{M-1}\}$ 
			range from $SD$ to $HD$.} and $t=8$ last time-steps:
		
		\subitem - last chunk bitrate (1 input):  
		\subitem $input_{0} = \frac{resolution_{1}}{resolution_{M-1}}$
		
		\subitem - current buffer size~\footnote{This variable, along 
			with the download 
			time variables, were normalized by $10$ seconds.} (1 input): 
		\subitem $input_{1} =  4$
		
		\subitem - throughput of the last $t$ chunks ($t$ 
		inputs): 
		\subitem $input_{2},...,input_{t+1} =  \frac{resolution}{download \
			time}$
		
		\subitem - download time of the last $t$ chunks 
		($t$ inputs): 
		\subitem $input_{t+2},...,input_{2t+1} = k$ for all~\footnote{With 
		discrete 
			jumps of $0.1$ seconds.} $k \in [4, 15]$
		
		\subitem - resolution options ($M$ inputs):  
		\subitem $input_{2t+2},...,input_{2t+M+1}=\{res_{0},...,res_{M-1}\}$	
		
		\subitem - remaining chunks~\footnote{In order to start with 
			all initial states for the fixed download time, we allow any 
			normalized 
			value of remaining chunks, thus, effectively encoding an infinite 
			amount of 
			initial states.}
		($1$ input):  
		\subitem $input_{2t+M+2}\in [0, 1]$	
		
		\subitem -suggested resolution probabilities ($M$ outputs): 
		\subitem 
		$max\{output_{0},...,output_{M-2}\}>output_{M-1}$~\footnote{$output_{M-1}=Pr(HD).$}

	\end{itemize}

	\subsection{\deepRM Properties}
	
	\begin{itemize}[leftmargin=*]
		\item \textbf{Property 1.} 
		For \textit{any} resource
		utilization level, when the queue is filled with identical jobs, the DRL
		system's output assigns a higher probability to
		\textit{schedule$_2$} than to \emph{wait}.
	\end{itemize}
	
	For $Q=5$ queued jobs of equal sizes (small or large) and for $d=10$ 
	memory 
	units, and $d'=10$ CPUs:
	
	\subitem - memory utilization inputs ($d$ inputs): 
	\subitem $input_{0},...,input_{d-1} \in [0, 1]$~\footnote{
		As mentioned, we abstracted the normalized utilization level. This 
		abstraction is encoded by bounding the utilization level in the range 
		$[0,1]$.}
	
	\subitem - CPU utilization inputs ($d'$ inputs): 
	\subitem $input_{d},...,input_{d+d'-1} \in [0, 1]$
	
	\subitem - suggested preference probabilities ($Q+1$ outputs):  
	\subitem $Pr($ \textit{schedule$_2$} $)= output_{1} > output_{Q} = Pr($ 
	\emph{wait} $)$ \\
	
	We note that in the DeepRM case study, some of the input variables were 
	encoded 
	by 
	multiple coordinates in the original input vector, as can be seen in the 
	verfication queries, available in our supplied 
	artifact~\cite{ArtifactGitRepo}.

\end{appendices}

\end{document}